\documentclass[conference]{IEEEtran}  
\usepackage{natbib}  
\usepackage{amsmath}    
\usepackage{graphicx}   
\usepackage{hyperref}   
\usepackage{amssymb}    
\usepackage{algorithm}  
\usepackage{algpseudocode} 
\usepackage{booktabs}
\usepackage{xcolor}
\usepackage{multirow}
\usepackage{subcaption} 
\usepackage{colortbl}
\usepackage{paralist}
\definecolor{highlight}{HTML}{BFFFBF}
\newcommand{\remove}[1]{} 
\newcommand{\param}{\Theta} 

\title{Architecture-Aware Learning Curve Extrapolation via Graph Ordinary Differential Equation}
 
\author{
\IEEEauthorblockN{First Author\IEEEauthorrefmark{1}, Second Author\IEEEauthorrefmark{2}}
\IEEEauthorblockA{\IEEEauthorrefmark{1}Department of XYZ, University A, City, Country \\
Email: first.author@example.com}
\IEEEauthorblockA{\IEEEauthorrefmark{2}Department of ABC, University B, City, Country \\
Email: second.author@example.com}
}

\author{ 
    Yanna Ding\textsuperscript{\rm 1}, 
    Zijie Huang\textsuperscript{\rm 2}, 
    Xiao Shou\textsuperscript{\rm 3},
    Yihang Guo\textsuperscript{\rm 2},
    Yizhou Sun\textsuperscript{\rm 2},
    Jianxi Gao\textsuperscript{\rm 1} \\
    \textsuperscript{\rm 1}Rensselaer Polytechnic Institute\\
    \textsuperscript{\rm 2}University of California, Los Angeles\\
    \textsuperscript{\rm 3}Baylor University 
}

\begin{document}
\maketitle

\begin{abstract} 
Learning curve extrapolation predicts neural network performance from early training epochs and has been applied to accelerate AutoML, facilitating hyperparameter tuning and neural architecture search. However, existing methods typically model the evolution of learning curves in isolation, neglecting the impact of neural network (NN) architectures, which influence the loss landscape and learning trajectories. In this work, we explore whether incorporating neural network architecture improves learning curve modeling and how to effectively integrate this architectural information. Motivated by the dynamical system view of optimization, we propose a novel architecture-aware neural differential equation model to forecast learning curves continuously. We empirically demonstrate its ability to capture the general trend of fluctuating learning curves while quantifying uncertainty through variational parameters. Our model outperforms current state-of-the-art learning curve extrapolation methods and pure time-series modeling approaches for both MLP and CNN-based learning curves. Additionally, we explore the applicability of our method in Neural Architecture Search scenarios, such as training configuration ranking. 
\end{abstract}
\section{Introduction \label{sec:intro}}
\begin{figure*}[th]
\centering
\includegraphics[width=0.9\textwidth]{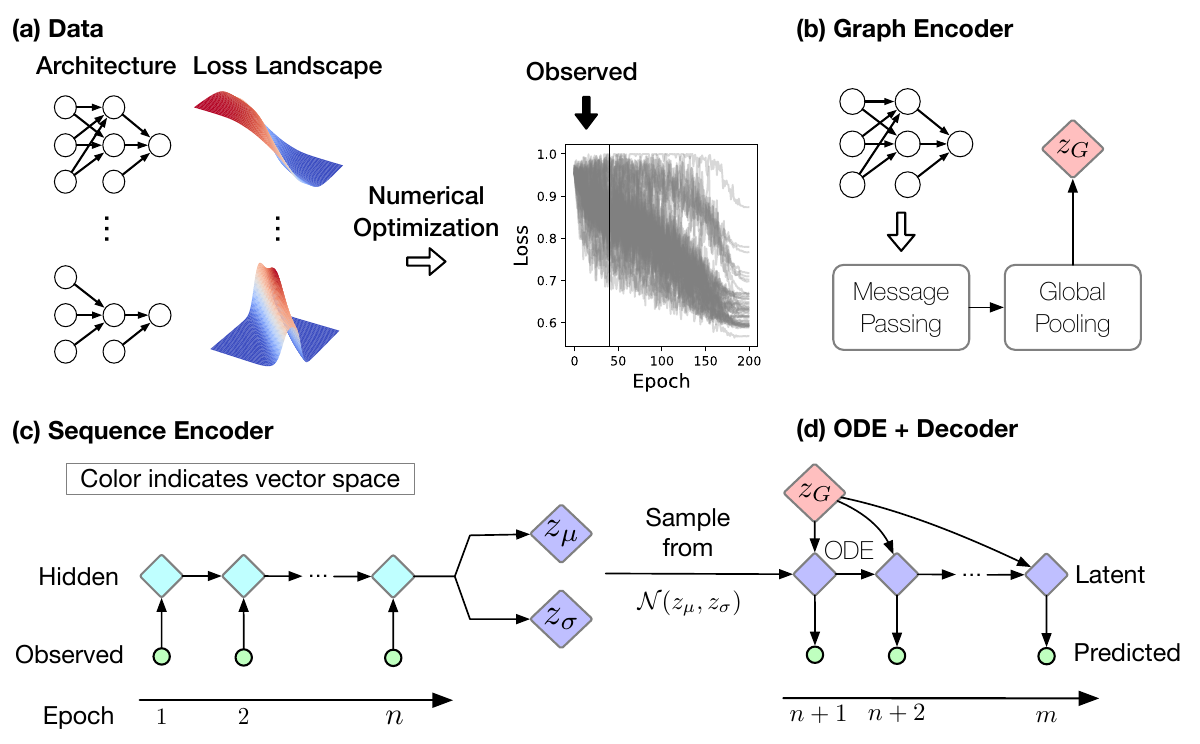}\caption{  
Overall framework. 
(a) Given fixed training data and a specific task, each architecture defines a unique loss landscape. We employ a numerical optimization method (e.g., gradient descent) to sample a loss trajectory across this landscape. Our dataset consists of various architectures paired with their corresponding loss curves. The model input is an observation window, which it utilizes to predict the trajectory within the subsequent prediction window.
(b) Our model architecture incorporates a graph encoder that captures the architectural structure by extracting a single embedding through message passing and global pooling.
(c) We initialize a latent distribution at the first epoch. A GRU unit processes this information, generating a hidden vector at each observed timestamp based on the prior hidden state and the current loss value.
(d) Finally, we integrate the ODE that governs the evolution of the latent loss states, with each time step modulated by the graph embedding.
}\label{fig:illus} 
\end{figure*}

Training neural architectures is a resource-intensive endeavor, often demanding considerable computational power and time. Researchers have developed various methodologies to predict the performance of neural networks early in the training process using learning curve data. Some methods~\citep{domhan2015speeding,gargiani2019probabilistic,adriaensen2024efficient} apply Bayesian inference to project these curves forward, while others employ time-series prediction techniques, such as LSTM networks. Despite their effectiveness, these approaches~\citep{swersky2014freeze, baker2017accelerating} typically overlook the architectural features of networks, missing out on crucial insights that could be derived from the models' topology. 

On another front, architecture-based predictive models have been developed to forecast network performance based purely on NN structures~\citep{shi2019multi,friede2019variational,wen2020neural,tang2020semi,yan2020does,ning2020generic,xu2019rnas,siems2020bench}. These models facilitate a deeper understanding of the relationship between architectures and their performance. However, they are limited in their ability to predict precise learning curve values at specific epochs and struggle to capture the variability in performance that a single architecture can exhibit under diverse training conditions.

Moreover, there is a growing interest in conceptualizing the optimization process during NN training as a dynamical system. By considering the step size in gradient descent as approaching zero, it is possible to formulate an ordinary differential equation for the model parameters \citep{su2016differential, zhang2023trained,maskan2024variational}. This perspective is useful for analyzing the convergence of different optimization algorithms, especially for convex problems.  Building on this foundational idea, we propose an innovative approach that models the evolution of learning curves using neural differential equations~\citep{chen2018neural}, tailored for a inductive setting where the trained learning curve predictor is applicable to new learning curves generated under various training configurations, such as different architectures, batch sizes, and learning rates. This approach leverages recent advancements in differential equations to provide a flexible framework capable of handling the complexities of modern neural training processes.

Our method merges the structural attributes of neural architectures with the dynamic nature of learning curves. We utilize a seq2seq variational autoencoder framework to analyze the initial stages of a learning curve and predict its future progression. This predictive capability is further enhanced by an architecture-aware component that produces a graph-level embedding from the architecture's topology, employing   techniques like Graph Convolutional Networks (GCN)~\citep{kipf2016semi} and Differentiable Pooling~\citep{ying2018hierarchical}. This integration not only improves the accuracy of learning curve extrapolations compared to existing methods  but also significantly  facilitates model ranking, potentially leading to more efficient use of computational resources, accelerated experimentation cycles, and faster progress in the field of machine learning.

Our contributions are twofold:
\begin{itemize}
\item We introduce an architecture-aware, dynamical system-based approach to model learning curves from different architectures for a given source task. Our model can predict the learning curves of unseen architectures using only a few observed epochs.
\item Our method improves model ranking by analyzing just a limited number of learning curve epochs, such as 10. This approach speeds up model selection by 20 times compared to traditional full-cycle stochastic gradient descent training.
\end{itemize}

\section{Related Work \label{sec:related-work}}

\paragraph{Learning curve extrapolation.}
Previous studies have explored learning curve prediction through diverse approaches~\citep{swersky2014freeze,domhan2015speeding,baker2017accelerating,chandrashekaran2017speeding,gargiani2019probabilistic,ru2021speedy,klein2022learning,adriaensen2024efficient}. A line of work has focused on Bayesian frameworks. Specifically, \cite{domhan2015speeding} utilized a weighted combination of functions to predict mean future validation accuracy and facilitate early termination of underperforming training runs. Building on this, \cite{chandrashekaran2017speeding} extended basis function extrapolation by incorporating historical learning curves from previous training runs, while \cite{klein2022learning} proposed a Bayesian neural network to flexibly model learning curves, removing the constraint that each epoch must outperform the previous one and thereby reducing instability in predictions. More recently, \cite{adriaensen2024efficient} applied a prior-data-fitted network training paradigm to enhance sampling efficiency from the posterior distribution of learning curves.  
Despite these advancements, existing methods overlook the role of architectural design, whereas our work explicitly incorporates this information to better model and understand the evolution of learning curves.

\textbf{Architecture-based performance prediction.}  
Advances in Graph Neural Networks (GNNs) have led to innovative methods for predicting the performance of neural network architectures~\citep{shi2019multi,friede2019variational,wen2020neural,tang2020semi,yan2020does,ning2020generic,xu2019rnas,siems2020bench}. In exploring unsupervised learning strategies, \cite{yan2020does} used architecture embeddings created by a pre-trained model to feed a Gaussian Process model for performance prediction. These studies often incorporate foundational GNN models such as the GCN \citep{kipf2016semi} and the Graph Isomorphism Network (GIN)~\citep{xu2018powerful} to effectively process input architectures. Various training objectives have been considered, including Mean Squared Error (MSE)~\citep{shi2019multi,wen2020neural,ning2020generic,tang2020semi}, graph reconstruction loss~\citep{friede2019variational,tang2020semi,yan2020does}, and pair-wise ranking loss~\citep{ning2020generic,xu2019rnas}, highlights the diverse methods aimed at improving the prediction of architecture performance. Building on these foundations, our approach goes beyond simply using the graph representation to extract a scalar performance value; instead, it integrates the graph information into the ODE of loss and can extrapolate to any time step of interest, including the value at convergence.

\paragraph{Dynamical system modeling.}
Neural ordinary differential equations (NODE)~\citep{chen2018neural} introduce a general framework for parameterizing ODEs with deep neural networks, deriving backpropagation through the adjoint sensitivity method. Variational autoencoders combined with NODE, as introduced in~\citep{rubanova2019latent}, are used to predict dynamics from irregularly sampled time-series data. Building on these developments, recent works~\citep{huang2020learning,huang2021coupled,luo2023hope,huang2024treat,CFGODE,cgode_www} further integrate graph neural networks (GNNs) with NODE to model temporal graphs, allowing for the representation of evolving nodes and edges over time. 
In contrast to modeling individual nodal trajectories, this work employs GNNs as a graph reduction technique, using their embeddings to drive the evolution of learning curves within the NODE framework.

\section{Method}

\subsection{Problem Formulation}
Learning curves, such as train or test loss, are generated by optimizing a neural network on various source tasks including adult income classification, image classification, and housing price regression~\citep{vanschoren2014openml}. Our goal is to train a single latent ODE model capable of extrapolating learning curves for a given source task across different architectures. The model infers full learning curves of length 
$m$ using only the initial 
$n$ epochs $y_1,\ldots,y_n$ and the corresponding network architecture, denoted as $G$.

Our approach transforms the conventional discrete optimization process into a continuous domain, operating within the continuous time interval $[0, T_{\max}]$, where $0$ marks the start and $T_{\max}$ corresponds to the last epoch $m$. The time for each epoch, $t_i$, is defined as $i \Delta t$ with $\Delta t = T_{\max}/m$.
 
We employ a seq2seq variational autoencoder framework~\citep{rubanova2019latent,huang2020learning,huang2021coupled}, where a sequence encoder parameterized by $ \phi $ processes the early part of the learning curve to estimate the variational parameters of the posterior distribution  $q_\phi$, which determines the latent state at the start of the prediction period, represented by $ \boldsymbol{z}_{n+1} \in \mathbb{R}^D$. A numerical solver then integrates an ODE function, denoted as $f :\mathbb{R}^D \rightarrow \mathbb{R}^D$, governing this latent state, starting from the initial condition $\boldsymbol{z}_{n+1}$. Each subsequent latent state $\boldsymbol{z}_i$ is decoded independently to predict the output $\hat{y}_{i}$ for $i > n$. 
This process is mathematically represented as follows:
\begin{align}
\boldsymbol{z}_{n+1} &\sim  q_{\phi} (\boldsymbol{z}_{n+1} | \{y_i,t_i\}_{i=1}^{n})\\
\boldsymbol{z}_{n+1}, \ldots, \boldsymbol{z}_{m} 
&= \mathrm{ODESolve}(f , \boldsymbol{z}_{n+1},  \nonumber\\
&\quad (t_{n+1}, \ldots, t_{m}), G)  \\
\hat{y}_i& =\mathrm{Decoder}(\boldsymbol{z}_i) \quad \text{for} \; i > n
\end{align}  
Here $\mathrm{ODESolve}(\cdot)$ simulates the ODE function $f(\cdot)$ to compute the latent states at time steps $t_{i}$ ($i\geq n+1$), given initial condition $\boldsymbol{z}_{n+1}$ and the neural architecture $G$. $\mathrm{Decoder}(\cdot)$ is a neural network mapping $\boldsymbol{z}_i$ to the corresponding output $\hat{y}_i.$ 
Since the model has access to only partial information about the optimization process, it cannot fully determine the loss landscape or accurately trace the trajectory within it. Therefore, we employ a variational framework to quantify uncertainty by estimating the most probable values and their variability. We refer to our method as \underline{L}earning \underline{C}urve \underline{G}raph\underline{ODE} (LC-GODE),  
highlighting the integration of architectures within the ODE framework to model learning curves. The illustration of our approach is shown in Figure~\ref{fig:illus}.

\subsection{Architecture-aware Differential Equation}

\paragraph{Observed time series encoder.}
We use a sequence encoder to compute the mean and standard deviation of the posterior distribution $q_{\phi}(\boldsymbol{z}_{n+1} | \{y_i,t_i\}_{i=1}^{n})$, which is assumed to be Gaussian:
\begin{align}
&\boldsymbol{z}_{n+1}  \sim q_{\phi}(\boldsymbol{z}_{n+1} | \{y_i,t_i\}_{i=1}^{n}) = \mathcal{N}(\mu_{\boldsymbol{z}_{n+1}}, \sigma_{\boldsymbol{z}_{n+1}}) \\
&\mu_{\boldsymbol{z}_{n+1}}, \sigma_{\boldsymbol{z}_{n+1}}  = \mathrm{SeqEncoder}_{\phi}(\{y_i, t_i\}_{i=1}^{n})
\end{align} 
We implement this using an RNN with GRU units for the sequence encoder. Other encoder options include Self-Attention~\citep{vaswani2017attention} and Temporal Convolutional Networks (TCN)~\citep{pandey2019tcnn}, which adapt well to varying observation lengths. We conduct an ablation study in the experimental section to evaluate the performance of these alternative sequence encoder implementations.

\paragraph{Architecture encoder.}
Assuming the graph representation $G$ contains $N$ nodes, the adjacency matrix $A \in \mathbb{R}_{\geq 0}^{N \times N}$ details node connections, where $A_{ij} > 0$ represents the edge from  node $i$ to node $j$.   We analyze two foundational neural network types: MLPs and CNNs. In MLPs, nodes correspond to neurons and edges correspond to the presence of the connections between neurons, whereas in CNNs, nodes represent feature maps and edges depict operations like $1\times1$ and $3\times3$ convolutions, or $3\times3$ average pooling. For CNNs, we adopt the cell-based representation~\citep{dong2020bench,liu2018darts}, which consists of four principal building blocks: stems, normal cells, reduction cells, and classification heads. The stem block is a fixed sequence of convolutional layers to process the input images. This is followed typically by 14-20 cells with reduction cells placed at 1/3 and 2/3 of the total depth. A normal cell contains 4 nodes, each of which belongs to the set of operations: skip connections, identity or zero (indicating the presence or absence of connections between certain layers), $1\times1$ and $3\times3$ convolutions, and $3\times3$ average pooling. Finally, the classification head employs a global pooling layer followed by a single fully connected layer and returns the network's output. Since the cell is repeated throughout this macro-skeleton, a CNN can be uniquely represented by its cell.

To derive a graph-level embedding, we first implement node-level message passing, followed by global pooling to extract a global embedding. Our method differs from existing architecture-based performance prediction approaches~\citep{liu2018darts,wen2020neural,knyazev2021parameter} by treating the node as a feature map rather than focusing on operations. Nodal features are calculated from the in-degree and out-degree of each node, normalized by the total number of edges. For MLPs, edge features are binary, while for CNN cells, they are integers representing operation types: $\{\text{zeroize}: 0, \text{$1\times 1$ conv}: 1, \text{$3\times 3$ conv}: 2, \text{$3\times 3$ avg pooling}: 3\}$. 
The nodal feature matrix is denoted as $X \in \mathbb{R}^{N \times d}$. The graph encoding process returns a vector representation of the architecture.
\begin{align}
\boldsymbol{z}_{G} = \mathrm{ArchEncoder}_{\theta_1} (X).
\end{align}
For node-level message passing, we employ GCN layers and normalize the adjacency matrix to stabilize training, following ~\citep{kipf2016semi}.
$$
\tilde{A} = A + I, \quad \tilde{D} = \sum_j \tilde{A}_{ij}.
$$ 
A GCN layer then transforms the nodal features into a hidden representation:
\begin{align}
Z = \tilde{D}^{-1/2} \tilde{A} \tilde{D}^{-1/2} X W
\end{align}
where $W \in \mathbb{R}^{D \times d}$ is the feature transformation matrix. Applying $l$ GCN layers aggregates information from $l$-hop neighbors. We employ learnable pooling, among other methods such as average- and max-pooling, and investigate each in our ablation study. The graph embedding $\boldsymbol{z}_{G}$ contributes to the evolution of the latent state, as detailed in the next section, which introduces the latent ODE.  
 
\paragraph{Latent ordinary differential equation.} The transformation from discrete to continuous domain is predicated on the assumption that the step size, or learning rate, approaches zero, thereby approximating the time derivative of NN parameters $\Theta$ using $\frac{d\Theta}{dt} = - \frac{\partial \mathcal{L}}{\partial \Theta}$~\citep{su2016differential}, where $\mathcal{L}$, $\Theta$ denotes the objective function and  parameters from the source task. Assuming the parameters implicitly depends on time $t$, the time derivative of the training loss can be written as:
\begin{align}
\frac{d{\mathcal{L} (\param(t))}}{d t} & = 
\frac{\partial \mathcal{L}}{\partial\Theta}^{\top} \frac{d \param }{dt} = -
\left(\frac{ \partial \mathcal{L}}{ \partial\param} \right)^\top\left(\frac{ \partial \mathcal{L}}{ \partial\param}\right)
\end{align}
The ground truth ODE for loss is independent of time.  
Therefore we adopt autonomous differential equations  to describe the continuous evolution of the learning curves. On the other hand, we do not directly use the exact formula, as this involves the computation of backpropagation of the source task, which is potentially computation intensive. Moreover, our primary goal is not to derive exact ODEs for each training configuration. Instead, we focus on efficiently inferring learning curves for new training configurations. To achieve this, we leverage the latent space, using the expressivity of hidden neurons to capture common patterns across learning curves from the same source task, despite variations in underlying architectures and hyperparameters.

Given that the training data for the source task remains fixed, the loss landscape varies depending on the architecture. Therefore, it's sufficient to model a universal latent ODE as a function of the architecture, enabling it to describe the evolution of various learning curves, each corresponding to a different architecture.
Our latent ODE is formalized by the equation: 
\begin{align}
\dot{\boldsymbol{z}} = f_{\theta_2}([\boldsymbol{z} || \boldsymbol{z}_{G}])~\label{eq:ode}
\end{align}
Here, $\dot{\boldsymbol{z}}$ denotes the derivative of the latent state vector $\boldsymbol{z}$ with respect to time, driven by the function $f_{\theta_{2}}$, which takes as input both the latent state $\boldsymbol{z}$ and a graph-level embedding $\boldsymbol{z}_{G}$. This combination allows the model to simultaneously consider the dynamic properties of the learning curve and the static characteristics of the architecture, enhancing the predictive capability of the system.  
Eq~\eqref{eq:ode} can be regarded as a single-agent representation of the dynamics induced by NN training, which involves multiple trajectories of neurons, edge weights, and loss. 
This reduced representation of coupled dynamical system has been explored in \citep{gao2016universal,laurence2019spectral} to study the tipping point of the original network dynamics. The difference from the prior dynamical system reduction approach is that the graph reduction mechanism is learnable so that the model can be adapted to unseen trajectories derived from different optimization trials.

Finally, numerical integration of Eq~\eqref{eq:ode} yields a time series of latent states $\boldsymbol{z}_i$ ($n<i\leq m$). Each latent state is independently decoded by a function  $\hat{y}_i = \mathrm{Decoder}_{\theta_3}(\boldsymbol{z}_i)$.

\paragraph{Training objective.}
To optimize our model, we  maximize the evidence lower bound (ELBO), fomulated as follows

\begin{align}
\mathrm{ELBO}(\phi,\theta) &= \mathbb{E}_{\boldsymbol{z}_{n+1} } [\log p_{\theta } (y_{n+1},\ldots,y_{m}  ) ]\nonumber \\
 &-  \mathrm{KL}[q_{\phi} (\boldsymbol{z}_{n+1} | \{y_i,t_i\})_{i=1}^{n} || p(\boldsymbol{z}_{n+1}) ]
\end{align} 
The first term in the ELBO equation represents the expected log-likelihood of observing future outputs given the latent states, as parameterized by $\theta=(\theta_1,\theta_2,\theta_3)$,  where $\theta_1$, $\theta_2$, and $\theta_3$ correspond to the architecture encoder, the ODE function, and the decoder, respectively.  The second term penalizes the divergence (measured by the KL-divergence) between the posterior distribution of the latent states and their prior distribution, enforcing a regularization that anchors the posterior closer to the prior. This balance ensures that while the model remains flexible enough to capture complex patterns in data, it also maintains a level of generalization that prevents overfitting.

\paragraph{Scalability and computational cost.}
The computational cost of numerical integration is minimal because the ODE models only the evolution of the loss embedding with dimension $D$, rather than modeling each node in the architecture individually. The runtime of the forward pass is bounded by $O(D^2T)$, where $T=m-n$ is the number of integration time steps, assuming the ODE function is implemented as an MLP with a fixed number of layers. Consequently, the runtime of the ODE component remains independent of the overall size of the neural network.

\begin{table*}[ht]
    \centering
    \small
    \caption{Extrapolation error for test accuracy curves derived from 2 tabular  tasks and 2 image classification  tasks. The percentage represents the fraction of the entire prediction window over which the error is computed and averaged. Green color highlights our approach. Bold denotes the best model and underline denotes the second best model. }\label{tab:extrapolation-test-acc} 
    
  \addtolength{\tabcolsep}{-0.15em}
    \begin{tabular}{l|cccccccccccc}
        \toprule
          & \multicolumn{6}{c}{car} & \multicolumn{6}{c}{segment}    \\
          &\multicolumn{3}{c}{MAPE}
          & \multicolumn{3}{c}{RMSE}
          &\multicolumn{3}{c}{MAPE} 
          &\multicolumn{3}{c}{RMSE}\\
        \cmidrule(lr){2-4} \cmidrule(lr){5-7}  \cmidrule(lr){8-10}  \cmidrule(lr){11-13}  
        Epochs & 80 & 140 &200& 80 & 140 &200& 80 & 140 &200& 80 & 140 &200\\
        \midrule 
         LC-BNN &0.5487 & 0.4887 & 0.4534 & 0.4232 & 0.3838 & 0.3592 & 0.6211 & 0.5688 & 0.5370 & 0.5384 & 0.4960 & 0.4694\\
         LC-PFN & \underline{0.0598}  & \underline{0.0681}  & \underline{0.0723}  & \underline{0.0443}  & 0.0502  & 0.0528 & \underline{0.0654}  & \underline{0.0708}  & \underline{0.0729}  & 0.0497  & \underline{0.0543}  & \underline{0.0555}   \\
         VRNN & 0.1857  & 0.1923  & 0.1923  & 0.1514  & 0.1511  & 0.1511 & 0.1840  & 0.1742  & 0.1742  & 0.1613  & 0.1557  & 0.1557 
 \\
         LSTM & 0.0853 & 0.1104 & 0.1251 & 0.0561 & 0.0709 & 0.0790 & 0.0825 & 0.1126 & 0.1346 & \underline{0.0478} & 0.0640 & 0.0758
  \\ 
        NODE & 0.0683  & 0.0730  & 0.0764  & 0.0459  & \underline{0.0499}  & 0.0531  & 0.0794  & 0.0837  & 0.0853  & 0.0574  & 0.0597  & 0.0609 
 \\
         NSDE & 0.0751  & 0.0768  & 0.0779  & 0.0503  & 0.0515  & \underline{0.0522}  & 0.0817  & 0.0854  & 0.0864  & 0.0595  & 0.0610  & 0.0614 
 \\ 
         \rowcolor{highlight} LC-GODE & \textbf{0.0431}  & \textbf{0.0463}  & \textbf{0.0488}  & \textbf{0.0328}  & \textbf{0.0349}  & \textbf{0.0365}  &\textbf{0.0566}  & \textbf{0.0591}  & \textbf{0.0608}  & \textbf{0.0462}  & \textbf{0.0477}  & \textbf{0.0487} 

  \\
       \midrule
         & \multicolumn{6}{c}{cifar10} & \multicolumn{6}{c}{cifar100}    \\
         &\multicolumn{3}{c}{MAPE}
         &\multicolumn{3}{c}{RMSE}
         &\multicolumn{3}{c}{MAPE} 
         &\multicolumn{3}{c}{RMSE}\\
        \cmidrule(lr){2-4} \cmidrule(lr){5-7}  \cmidrule(lr){8-10}  \cmidrule(lr){11-13}  
        Epochs & 80 & 140 &200& 80 & 140 &200& 80 & 140 &200& 80 & 140 &200\\
        \midrule
         LC-BNN & 0.4235 & 0.4204 & 0.4036 & 0.2101 & 0.2116 & 0.2260 & 1.2441 & 1.4087 & 1.2946 & 0.1161 & 0.1219 & 0.1373\\
         LC-PFN&  0.1514  & 0.1618  & 0.1782  & 0.0835  & 0.1023  & 0.1325 & \underline{0.3115}  & \underline{0.3536}  & \underline{0.3749}  & 0.0738  & 0.1062  & 0.1549 
 \\
         VRNN & \underline{0.1231}  & \underline{0.1316}  & 0.1419  & 0.0905  & 0.0917  & 0.0987 & 0.3621  & 0.5105  & 0.5801  & 0.1153  & 0.1161  & 0.1224    \\
         LSTM & 0.1391 & 0.1467 & 0.1309 & 0.0752 & 0.0736 & 0.0682 & 0.3853 & 0.5599 & 0.5761 & 0.0659 & 0.0715 & 0.0720\\ 
         NODE & 0.1457  & 0.1525  & 0.1404  & 0.0711  & 0.0717  & 0.0727 &0.4895  & 0.6593  & 0.6592  & 0.0681  & 0.0760  & 0.0798   \\
         NSDE & 0.1491  & 0.1454  & \underline{0.1245}  & \underline{0.0686}  & \underline{0.0670}  & \underline{0.0645}  & 0.4547  & 0.5076  & 0.4201  & \underline{0.0629}  & \underline{0.0669}  & \underline{0.0665}  \\   
        \rowcolor{highlight}LC-GODE &\textbf{0.1107}  & \textbf{0.1093}  & \textbf{0.0913}  & \textbf{0.0645}  & \textbf{0.0603}  & \textbf{0.0557}  &\textbf{0.2708}  & \textbf{0.3205}  & \textbf{0.2661}  & \textbf{0.0621}  & \textbf{0.0636}  & \textbf{0.0625} 

  \\ 
        \bottomrule
    \end{tabular}
\end{table*}

\section{Experiments}
\label{sec:exp}

We demonstrate the capability of LC-GODE to forecast model performance across a range of common AutoML benchmarks. First, we evaluate LC-GODE against six learning curve extrapolation techniques on real-world datasets obtained using stochastic gradient descent on both tabular and image tasks. The training process is conducted individually for each source task.
Furthermore, we assess our method’s effectiveness in ranking training configurations by predicted optimal performance.
Lastly, we explore model sensitivity to variants of architecture, observed time-series encoders and  hyper-parameters.
Our code and supplemental materials are publicly available\footnote{\url{https://github.com/dingyanna/LC-GODE.git}}.

\paragraph{Datasets.} 
We consider test loss and test accuracy curves for both MLP-based and CNN-based architectures. Specifically for MLPs, we use \verb|car| and \verb|segment| tabular data binary classification tasks from OpenML~\citep{vanschoren2014openml} as source tasks. Following LCBench~\citep{zimmer2021auto}, we randomly generate training configurations that include variables such as the number of layers, number of hidden units, and learning rates. For each dataset, we conduct 550 optimization trials across 200 epochs. For CNN-based models, we employ the NAS-Bench-201 dataset~\citep{dong2020bench}, which provides comprehensive learning curves for each architecture over a span of 200 epochs across two image datasets: CIFAR-10, CIFAR-100~\citep{krizhevsky2009learning}. We randomly select 5,000 architectures from CIFAR-10 and CIFAR-100 to form our dataset. Furthermore, we reserve 20\% of all trials as the test set for each MLP source task and 25\% for each CNN source task.

\subsection{Extrapolating Real-world Learning Curves \label{exp:extrapolation}}

\paragraph{Experimental setup.}
The goal of this experiment is to evaluate the LC-GODE model against established learning curve prediction methods using real-world benchmarks. We train our model separately on the test loss curves of each source task.  
The condition length is set to 10 epochs for all methods in this experiment.
The instantiation of LC-GODE that we report features: (i) an architecture encoder that utilizes {2} layers and employs a learnable pooling technique, (ii) an observed time-series encoder implemented using GRU, (iii) an ODE function with a 2-layer MLP and integrated using the Runge-Kutta 4 method~\citep{butcher1996history}.  Further details on the evaluation metrics and the training settings can be found in the supplemental materials.

\paragraph{Baselines.} 
We evaluate our model against six methods, including three Bayesian approaches and {three} general time-series prediction approaches.
\begin{inparaenum}[(i)]
\item LC-BNN: This method utilizes a Bayesian neural network to model the posterior distribution of future learning curves. The probability function is constructed from a combination of basis functions, following the approach described by \cite{domhan2015speeding}.
\item LC-PFN: A transformer-based model is trained on synthetic curves that are generated from a pre-defined prior. This method serves as an efficient alternative to traditional Markov Chain Monte-Carlo (MCMC) techniques for sampling from posterior distributions.
\item VRNN: A probabilistic model utilizing random forests and Bayesian recurrent neural networks.
\item LSTM: This Recurrent Neural Network captures sequential dependencies. Before the observation cutoff at $n$, input states are derived from actual observations. Post $n$, the model uses its own predictions from previous timestamps as inputs.
\item NODE: Latent Ordinary Differential Equations, focusing primarily on modeling the latent loss representation without incorporating architectural information.  
\item NSDE: Latent Stochastic Differential Equations, consisting of a drift term and a diffusion term. The drift term is the same as NODE, and the sequence encoder and decoder is the same as in our model.  
\end{inparaenum} 
Both NODE and NSDE are trained using the variational framework.

\begin{figure}[h]
  \centering
  \begin{subfigure}[b]{\columnwidth}
        \centering
        \includegraphics[width=\columnwidth]{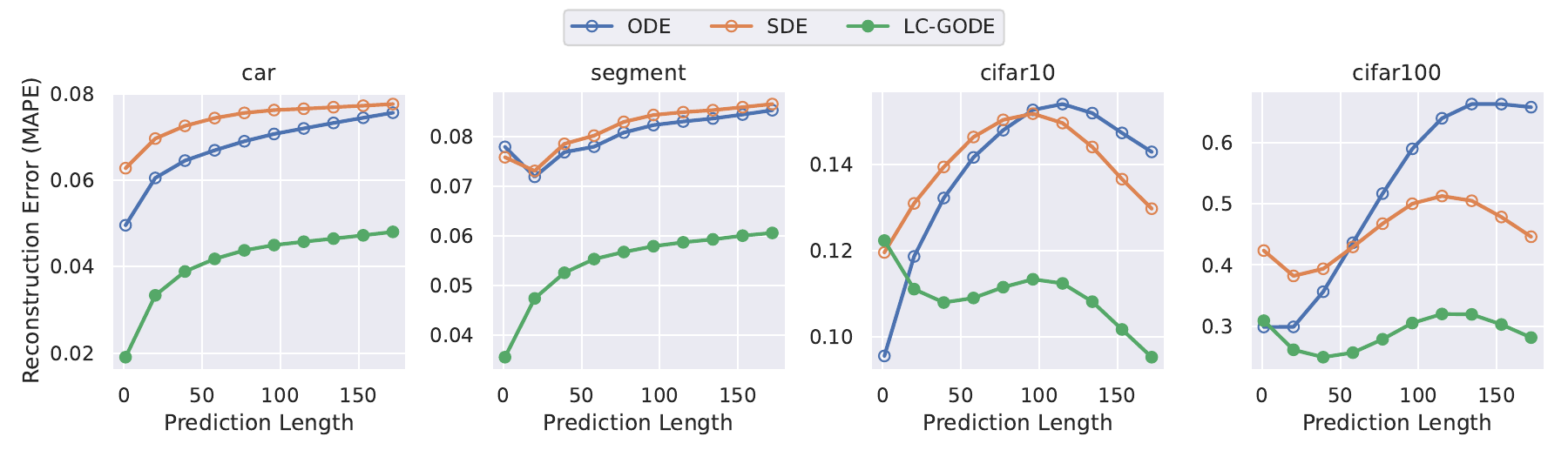}
        \caption{Test Accuracy}
        \label{fig:acc-extrap}
    \end{subfigure} 
    \begin{subfigure}[b]{\columnwidth}
        \centering
        \includegraphics[width=\columnwidth]{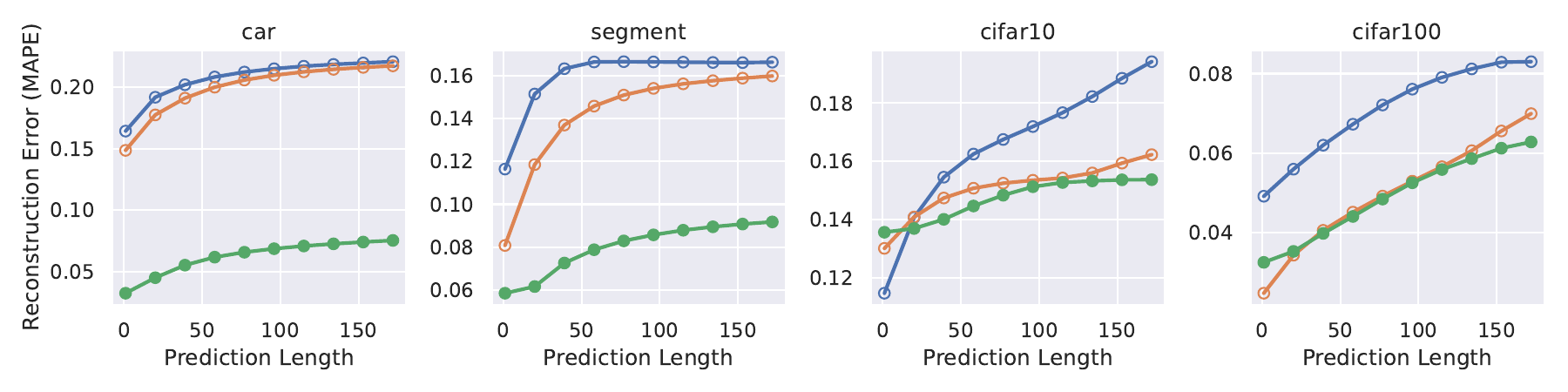}
        \caption{Test Loss}
        \label{fig:loss-extrap}
    \end{subfigure} 
    \caption{ Extrapolation error (MAPE) for learning curves w.r.t. prediction length. }\label{fig:extrap}
\end{figure}

\paragraph{Results.} 
\begin{figure}[h] 
\includegraphics[width=0.5\textwidth]{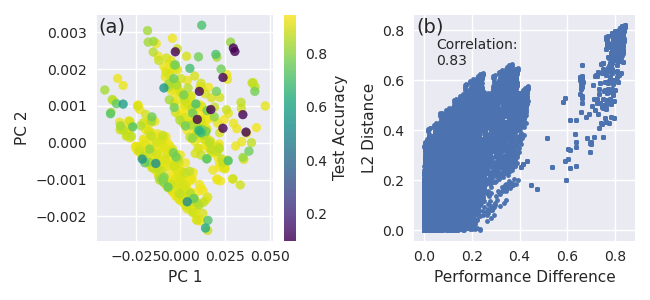}\caption{ (a) Graph embedding projections. (b) Pairwise initial latent embedding distance vs performance difference. } \label{fig:graph-emb}
\end{figure}
We evaluate performance using Mean Absolute Percentage Error (MAPE) and Root Mean Squared Error (RMSE) with different prediction lengths. Table~\ref{tab:extrapolation-test-acc} shows the extrapolation error for test accuracy curves from four datasets. Due to space constraints, the comparison results for loss curve extrapolation are provided in the supplement. Our proposed LC-GODE outperforms all baselines on these datasets. Specifically, LC-GODE reduces the error on test accuracy curves by 36.13\%, 30.72\%, 34.97\%, and 59.63\%, and on test loss curves by 65.5\%, 44.61\%, 20.1\%, and 23.45\% compared to the NODE model without architecture information. This improvement is due to the incorporation of architecture information with graph embedding. The architectures with similar performance are mapped to nearby locations in the hidden space, as shown in Figure~\ref{fig:graph-emb}(a). To demonstrate the advantage of jointly training the time-series encoder with the graph encoder, we compare the optimal performance difference of two configurations with the distance between their initial latent states in Figure~\ref{fig:graph-emb}(b). When architecture information is included, the correlation between these distances increases from 0.77 to 0.83 for the CIFAR-10 test accuracy curves. For a detailed comparison over epochs, we plot MAPE at 10 prediction lengths for NODE, NSDE, and LC-GODE on both test accuracy and loss curves (Figure~\ref{fig:extrap}).   
Overall, models perform better and have larger improvement on test accuracy curves compared to loss curves. This could be attributed to the fact that classification accuracy is less sensitive to decision boundary changes than cross-entropy loss, making it less variable and easier to predict.

Regarding the baseline comparisons, LC-PFN emerges as the second most effective approach for extrapolating test accuracy curves of the car and segment source tasks, while the NSDE model ranks as the second in predicting test loss curves for CIFAR-10 and CIFAR-100. Both NSDE and NODE models demonstrate comparable performance, with NSDE marginally outperforming NODE. These results underscore the viability of employing time-series approaches for addressing learning curve extrapolation challenges. Notably, the slight advantage of NSDE over NODE suggests subtle benefits in capturing stochastic dynamics that may be present in complex learning scenarios. This comparison highlights the potential for refined time-series models to enhance predictive accuracy and adaptability in diverse training environments.

\paragraph{Model ranking\label{exp:model-ranking}}

\begin{table*}[h]
    \centering 
     
    \caption{Model selection evaluation based on \textit{regret} (difference between the performance of the predicted best model and actual best model) and the \textit{ranking} of the predicted best model. The number in the bracket after the dataset denotes the total number of configurations.  } \label{tab:nas} 
     {
    \begin{tabular}{ll|ccc|ccc}
        \toprule
      \multirow{2}{*}{Metric}  & \multirow{2}{*}{Dataset}& \multicolumn{3}{c|}{Accuracy} & \multicolumn{3}{c}{Loss}  \\ 
      \cmidrule{3-5}\cmidrule{6-8}
      & & NODE & NSDE & LC-GODE& NODE & NSDE & LC-GODE \\
         \midrule
        \multirow{4}{*}{\textit{regret}}
         & car & 0.0023 & 0.0023  &  0.0023 &0.4570 &0.0950 &0.0000 \\
         &segment& 0.0052 & 0.0017  &0.0017 &0.0100 &0.0100 & 0.0100 \\
         &cifar10&0.0025&0.0101 & 0.0004 &0.0164 &0.0243 & 0.0048\\ 
         &cifar100 &0.0000  &0.0374 & 0.0000 &0.0262 &0.0606 &0.0303 \\
         \midrule
     \multirow{4}{*}{\textit{ranking}}
          &car (110)       &2 & 2      &2  &43 &4  & 1 \\
          &segment (110)   &18 & 7     &4  &2  &2  & 2\\
          &cifar10 (1250)  &10 & 86    &2  &21 &73 & 3  \\  
          &cifar100 (1250) &1 & 142    & 1  &87 &607& 137   \\
        \bottomrule
    \end{tabular}
    } 
\end{table*}

We evaluate our method's efficacy in ranking training configurations by comparing the predicted and true optimal performances at a fixed snapshot, employing two metrics: \textit{regret} (the performance discrepancy between the actual and predicted best configurations) and \textit{ranking} (the position of the predicted best configuration according to the true learning curves). These metrics demonstrate whether the predicted performance can effectively guide the selection of a performant configuration.

As shown in Table~\ref{tab:nas}, our proposed approach, LC-GODE, enhances the \textit{ranking} of the predicted best model by 3 positions on the segment dataset and by 8 positions on the CIFAR-10 dataset, and reduces \textit{regret} by 96\% on CIFAR-10 compared to the superior baseline among NODE and NSDE for test accuracy curves. Nevertheless, when employing test loss to identify the best model, the overall \textit{ranking} for all methods declines, indicating that predicted test loss is less effective for model selection.

Our approach identified a performant model with a 20x speedup compared to exhaustive brute-force training using stochastic gradient descent (SGD) on the MLP datasets. The speedup is computed as the total runtime needed to fully train all configurations via SGD, divided by the sum of the actual training time for $n$ epochs and the inference time required by LC-GODE. The inference time for LC-GODE is approximately 0.8 seconds, and the majority of the model selection time is spent on the initial $n$ epochs of SGD training. Additional results on the training and inference runtime of all methods are provided in the supplement.
 
Figure~\ref{fig:scatter} further shows the true vs predicted best test accuracy for the 2 tabular data and 2 image classification source tasks, further indicating a high correlation between predicted and true metrics.
\begin{figure}[h]
  \centering
  \includegraphics[width=\linewidth]{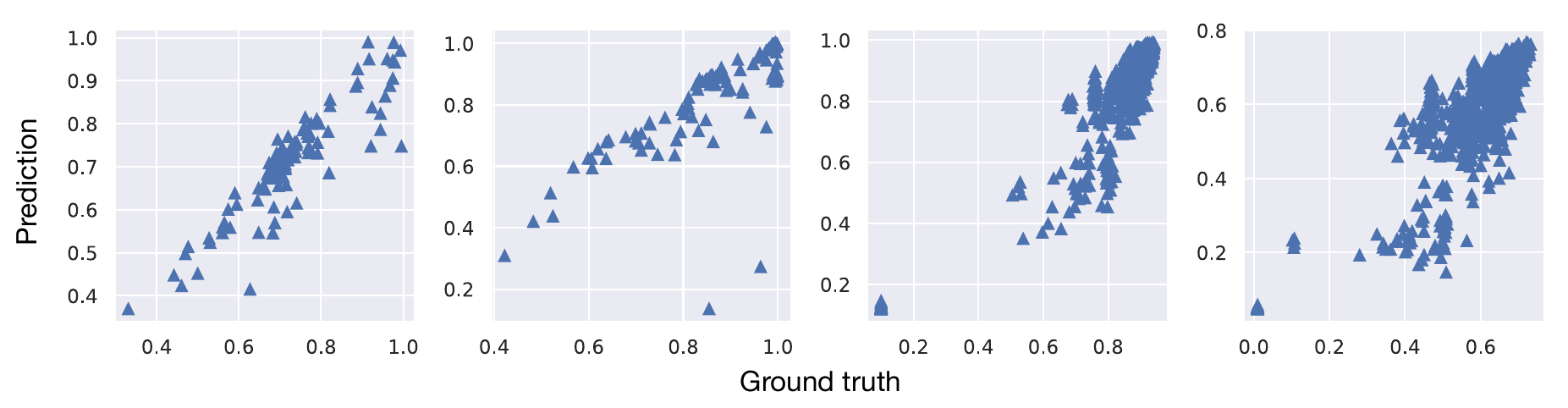}
  \caption{ True vs predicted best test accuracy for car, segment, CIFAR-10, CIFAR-100, respectively.  }  \label{fig:scatter} 
\end{figure}

\subsection{Ablation Study} 
\label{sec:ablation}
\begin{figure}[h]
  \centering
  \begin{subfigure}[b]{0.33\columnwidth}
        \centering
        \includegraphics[width=\textwidth]{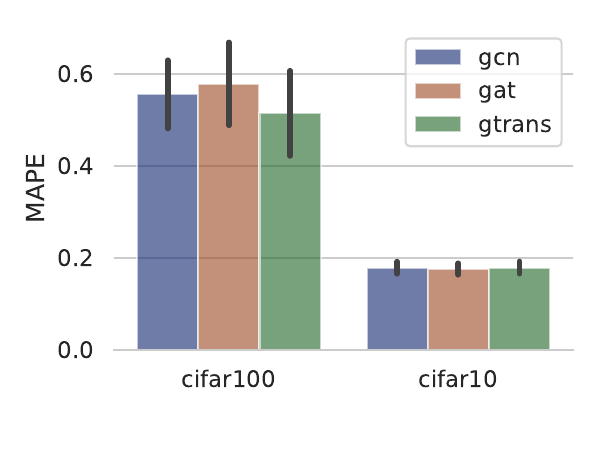}
        \caption{Message passing }
        \label{fig:ablation-gnn}
    \end{subfigure}%
    \begin{subfigure}[b]{0.33\columnwidth}
        \centering
        \includegraphics[width=\textwidth]{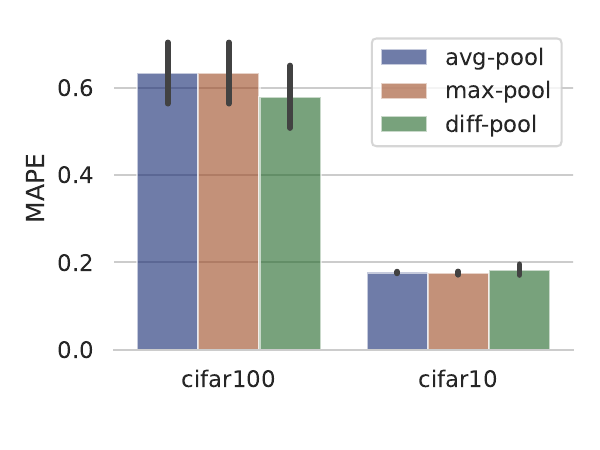}
        \caption{Graph pooling}
        \label{fig:ablation-pooling}
    \end{subfigure}%
     \begin{subfigure}[b]{0.33\columnwidth}
        \centering
        \includegraphics[width=\textwidth]{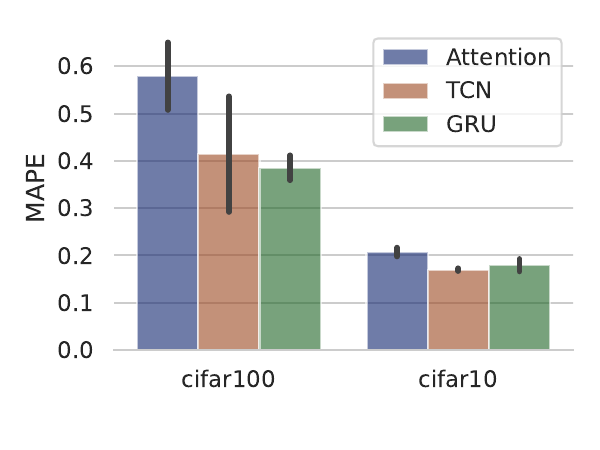}
        \caption{Sequence encoder}
        \label{fig:ablation-seq-enc}
    \end{subfigure} 
    \caption{Ablation study for encoders. }\label{fig:ablation}
\end{figure}

\begin{figure}[h]
  \centering 
    \centering
    \includegraphics[width=0.9\columnwidth]{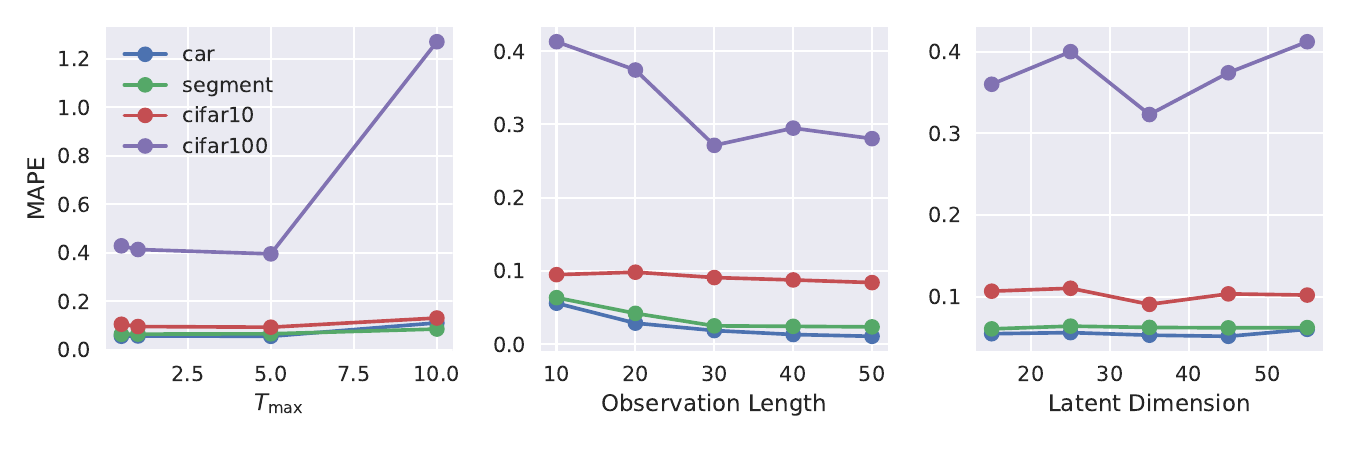}\caption{ Hyperparameter sensitivity study.  }  \label{fig:hyper} 
\end{figure}

We explore three variations for each of the following components: the message passing mechanism, graph pooling, and time series encoder. The message passing mechanisms include GCN~\citep{kipf2016semi}, Graph Attention Networks (GAT)~\citep{velivckovic2017graph}, and Graph Transformers~\citep{hu2020heterogeneous}. For pooling methods, we use average pooling, max-pooling, and a learnable pooling method based on DiffPool~\citep{ying2018hierarchical}, where a separate GCN module determines each node's contribution to the global embedding. The time-series encoder variations include self-attention~\citep{vaswani2017attention}, TCN~\citep{pandey2019tcnn}, and an autoregressive version of GRU~\citep{dey2017gate,kipf2018neural}. The model is trained using early stopping if no improvement is observed after 50 epochs.

As shown in Figure~\ref{fig:ablation}, for CIFAR-100, the combination of graph transformers, DiffPool, and GRU achieved the best results, while for CIFAR-10, GAT, max-pooling, and TCN performed better. This suggests that different combinations of encoders can affect extrapolation accuracy, with a more significant impact observed on CIFAR-100.

\paragraph{Hyperparameter sensitivity.} We analyze the impact of three hyperparameters: maximal time $T_{\max}$, observation length, and hidden dimension (Figure~\ref{fig:hyper}). A larger $T_{\max}$ negatively impacts performance. As observation length increases, the model receives more information about the curve, requiring less extrapolation and thus reducing error. The performance remains relatively constant when the latent dimension is within the range of 20 to 50.  


\section{Conclusion}
In this study, we introduced LC-GODE, a novel approach that merges architectural insights with learning curve extrapolation from a dynamical systems perspective. Our model uses early performance data to predict future learning curve trajectories, significantly enhancing predictions of both test loss and accuracy. This architecture-aware, dynamical system-based method not only surpasses existing extrapolation techniques but also enhances model ranking and the selection of optimal configurations by analyzing just a small number of epochs. This efficient approach achieves a 20x speedup in model selection compared to traditional full training cycles using stochastic gradient descent.  Future work could improve by incorporating the impact of source data to generalize across source tasks.

\section{Acknowledgement}
Y.N.D. and J.X.G. are supported by the National Science Foundation (No. 2047488), and by the Rensselaer-IBM AI Research Collaboration.
This work was partially supported by NSF 2211557, NSF 2119643, NSF 2303037, NSF 2312501, NASA, SRC JUMP 2.0 Center, Amazon Research Awards, and Snapchat Gifts.
\bibliographystyle{IEEEtranN}
\bibliography{reference} 
\newpage

\onecolumn
\appendices
\section{Experimental Details\label{supp:experiment-details}}

\paragraph{Datasets.}
To generate MLP learning curves, we selected 2 tabular data classification tasks from OpenML~\citep{vanschoren2014openml} as detailed in Table~\ref{tab:tabular-data}. We adhere to the standard procedure outlined in LCBench~\citep{zimmer2021auto}, with the exception that we introduce variability by randomly sampling the number of hidden units for each layer, rather than sampling only the maximal number of hidden units. All optimization runs are performed using SGD with decaying learning rate by 0.5 every 80 epochs. The total number of epochs is 200. We discard runs The parameters randomly sampled include four floating-point values and $2 + l$ integers, where $l$ denotes the number of layers.

\begin{table}[ht]
\caption{Hyperparameter setting used to generate MLP data. }\label{tab:mlp-hyper}
\centering
\begin{tabular}{lcc}
\toprule
\textbf{Hyperparameter }  & \textbf{Value} & \textbf{Log-scale} \\
\midrule 
Batch size &$[16, 512]$ & Y\\
Learning rate &$[0.0001, 0.1]$ & Y \\
Batch size & $[16, 512]$ & Y\\
Weight decay &  $[0.00001,0.1]$ & N\\
Number of layers &  $[1, 5]$  & N\\
Number of units per layer &  $[16, 1024]$ & Y \\
 Dropout & $[0.0, 1.0]$ & N \\ 
\bottomrule
\end{tabular}
\end{table}  

\begin{table*}[!ht]
    \centering
    \caption{Tabular datasets used to generate MLP learning curves. }\label{tab:tabular-data}
    \begin{tabular}{lcccccc}
    \toprule
        \textbf{Tabular dataset} & \textbf{\# train samples} & \textbf{\# test samples} & \textbf{\# features} & \textbf{\# labels}   \\ \midrule 
        car & 1296 & 432 & 6 & 2   \\    
        segment & 1732 & 578 & 19 & 2  \\    
        \bottomrule
    \end{tabular}
\end{table*}

\paragraph{Metrics.}
We adopt six metrics to evaluate our approach from three dimensions.
\begin{itemize}
	\item Trajectory reconstruction. To evaluate curve reconstruction error, we utilize MAPE. 
      Let $N_{pred}$ denote the prediction length. 
      Let $ \hat{x}^{(i)} \in \mathbb{R}^{N_{pred}} $ denote the $i$th trajectory in the test set.  Let $y^{(i)}\in \mathbb{R}^{N_{pred}}$ denote the ground truth trajectory. The reconstruction error for one trajectory is comuted as
		\begin{align}  \text{MAPE}_i = \frac{1}{N_{pred}}\sum_{t=1}^{N_{pred}}\left\lvert \frac{y_t^{(i)} - \hat{x}_t^{(i)}}{y_t^{(i)}} \right\rvert 
		\end{align}
    The error metric for the entire test dataset is the corresponding average over all trajectories.
    \begin{align} 
        \text{MAPE} = \frac{1}{N_{traj}} \text{MAPE}_i
    \end{align}
	\item Model Selection. We use Pearson correlation as adopted in [?] and Kendall $\tau$ [NasWOT].
	\item Efficiency. We report the training runtime per epoch and the the wall-clock time to perform inference for one architecture using our model. The speedup  is computed as
	\begin{align}
	\text{Speedup} = \frac{\text{Runtime for SGD over 52 Epochs}}{\text{Runtime for SGD over }T_{cond} \text{ epochs} + \text{LC-ODE Forward Runtime}}
	\end{align}
\end{itemize}

\paragraph{Hyperparameter Setting.}
Without further specification, the hyperparameters to train both our model and the baselines are set according to  Table~\ref{tab:hyper}
\begin{table}[ht]
\caption{Hyperparameter setting used throughout the paper. }\label{tab:hyper}
\centering
\begin{tabular}{lcc}
\toprule
\textbf{Hyperparameter }  & \textbf{Value} \\
\midrule
Latent Dimension & 16    \\
Learning Rate    & 0.001 \\ 
Batch Size       & 128 (CNN) 40 (MLP)    \\
Optimizer        & AdamW~\citep{loshchilov2017decoupled}\\
Number of Epochs & 400\\
Condition Length & 20\% $T_{max}$ \\
Prediction Length & 80\% $T_{max}$ \\
\bottomrule
\end{tabular}
\end{table}

\subsection{Baseline Configurations.}
\paragraph{LC-BNN.} 
LC-BNN is a function that maps a tuple containing a configuration and an epoch \((\text{configuration}, \text{epoch})\) to the loss value associated with that configuration at the specified epoch. We represent the trajectory data as pairs of input and target values:
$((\text{config}_i, t_j), y_i(t_j))$
where $\text{config}_i$ denotes the $i$th configuration, $t_j$ represents an epoch, and $y_i(t_j)$ is the corresponding metric value. 
For MLP configurations, we use the hyperparameters specified in LCBench, detailed in Table~\ref{tab:mlp-hyper}. In the case of NAS-Bench-201, we utilize the hyperparameters outlined in their respective publication. Note that all architectures within NAS-Bench-201 utilize a uniform set of hyperparameters. Nevertheless, we include these hyperparameters as input to LC-BNN, as it requires at least one additional input beyond the epoch number.
 
Our dataset is structured within an inductive learning framework, comprising multiple training trajectories, with the test set including unseen trajectories. Since LC-BNN does not have an associated conditional length, to ensure a fair comparison, we incorporate the conditional window from our test set into the training data for LC-BNN.

\paragraph{LC-PFN.} 
To apply LC-PFN, we preprocess our data using the normalization procedure outlined in Appendix A by~\cite{adriaensen2024efficient}. This approach ensures that our data is consistently formatted and scaled according to the specified guidelines.
The normalization process transforms observed curves into a constrained range $[0, u]$ where $u \approx 1$. LC-PFN takes as input the normalized observations and outputs inferred loss values. These values are then mapped back to their original space using the inverse of the normalization function. The parameters of the normalization function are defined as $\boldsymbol{\lambda} = (\text{minimize}, l_{\text{hard}}, u_{\text{hard}}, l_{\text{soft}}, u_{\text{soft}})$:
\begin{itemize}
	\item \textbf{min?}: A Boolean indicating whether the curve is to be minimized or maximized.
	\item $l_{\text{hard}}, u_{\text{hard}}$: Hard bounds defining the absolute limits of the learning curve.
	\item $l_{\text{soft}}, u_{\text{soft}}$: Soft bounds that guide the behavior of the learning curve.
\end{itemize} 
The normalization function is formulated as:
\begin{align*}
g_{\boldsymbol{\lambda}}(x) = \text{cr}_{0.5}\left(\frac{c}{1 + e^{-(a(x - b))}  } + d\right) 
\end{align*}
where: 
\begin{align*}
cr_{0.5}(y) &=
\begin{cases}
1- y \quad &\text{if \textbf{min?}}\\
y \quad &\text{otherwise}
\end{cases}\\
a &= \frac{2}{u_{\text{soft}} - l_{\text{soft}}}, \quad b = -\frac{u_{\text{soft}} + l_{\text{soft}}}{u_{\text{soft}} - l_{\text{soft}}}, \\
c &= \frac{1 + \exp(-a(u_{\text{hard}} - b)) + \exp(-a(l_{\text{hard}} - b)) + \exp(-a(u_{\text{hard}} + l_{\text{hard}} - 2b))}{\exp(-a(l_{\text{hard}} - b)) - \exp(-a(u_{\text{hard}} - b))}, \\
d &=  \frac{c}{1 + \exp(-a(l_{\text{hard}} - b))}.
\end{align*}
The inverse of $g^{-1}_{\boldsymbol{\lambda}}(y) $ is defined as
\begin{align}
g^{-1}_{\boldsymbol{\lambda}}(y) =b - {\log \left(\frac{c}{\text{cr}_{0.5}(y)-d}-1\right)}/{a}
\end{align}
For normalizing observed log loss values, we utilize $\boldsymbol{\lambda} = (\text{True}, 0, \log(10), 0, \max\{\mathbf{y}_{0}^{train}\})$. Here $\mathbf{y}_{0}^{train}$ is the log loss value at the first epoch of the trajectories to train AutoML models. To normalize accuracy curves, we apply $\boldsymbol{\lambda} = (\text{False}, 0, 1, 0, 1)$.

\paragraph{LSTM.} 
The LSTM implementation is adapted from the published code associated with NRI~\citep{kipf2018neural}. This implementation employs teacher forcing by utilizing the observation window. Specifically, it features a \verb|step()| module, which comprises a Long-short-term memory (LSTM) block and a two-layer MLP with ReLU activation. The \verb|step()| function processes the previous input state and hidden state, outputting the prediction and hidden state for the next immediate time step. The \verb|forward()| function iteratively calls \verb|step()| for a total of $T_{max}$ times. During the initial $T_{cond}$ steps, the observed learning curve data is used as the input state. For the subsequent steps beyond $T_{cond}$, the output from the previous prediction is used as the new input state.

\section{Additional Results\label{sec:additional}}

\begin{table*}[ht]
    \centering
    \small
    \caption{Extrapolation error for test loss curves derived from 2 tabular  tasks and 2 image classification tasks computed over three prediction lengths, observing 10 epochs.     }\label{tab:extrapolate-loss}
    
  \addtolength{\tabcolsep}{-0.15em}
    \begin{tabular}{l|cccccccccccc}
        \toprule
          & \multicolumn{6}{c}{car} & \multicolumn{6}{c}{segment}    \\
          &\multicolumn{3}{c}{MAPE}
          & \multicolumn{3}{c}{RMSE}
          &\multicolumn{3}{c}{MAPE} 
          &\multicolumn{3}{c}{RMSE}\\
        \cmidrule(lr){2-4} \cmidrule(lr){5-7}  \cmidrule(lr){8-10}  \cmidrule(lr){11-13}  
        Epochs & 80 & 140 &200& 80 & 140 &200& 80 & 140 &200& 80 & 140 &200\\
        \midrule 
        LC-BNN & 0.4545 & 0.3715 & 0.3295 & 0.2798 & 0.2348 & 0.2096 & 0.4322 & 0.3893 & 0.3785 & 0.2358 & 0.2036 & 0.1877
\\
         LC-PFN & \underline{0.0723}  & \underline{0.0906}  & \underline{0.0999}  & \underline{0.0349}  & \underline{0.0397}  & \underline{0.0425} &\underline{0.0795}  & \textbf{0.0890}  & \underline{0.0937}  & \textbf{0.0329}  & \textbf{0.0348}  & \textbf{0.0361 }
 \\
         VRNN & 0.2216  & 0.2054  & 0.2054  & 0.1306  & 0.1281  & 0.1281 & 0.3308  & 0.3660  & 0.3660  & 0.1518  & 0.1660  & 0.1660 
 \\
         LSTM & 0.1104 & 0.1517 & 0.1733 & 0.0539 & 0.0697 & 0.0774 & \textbf{0.0782} & 0.0953 & 0.1076 & 0.0343 & 0.0386 & 0.0429 \\ 
         NODE &0.2110  & 0.2183  & 0.2218  & 0.0909  & 0.0927  & 0.0939  & 0.1665  & 0.1662  & 0.1670  & 0.0623  & 0.0635  & 0.0649 
 \\
         NSDE &0.2039  & 0.2141  & 0.2183  & 0.0884  & 0.0900  & 0.0906  & 0.1493  & 0.1574  & 0.1607  & 0.0558  & 0.0581  & 0.0593    \\
           LC-GODE &  \textbf{0.0644}  & \textbf{0.0722}  & \textbf{0.0765}  & \textbf{0.0292}  & \textbf{0.0307}  & \textbf{0.0317}  & {0.0816}  & \underline{0.0892}  & \textbf{0.0925}  & \underline{0.0342}  & \underline{0.0368}  & \underline{0.0378} 
 \\
       \midrule
         & \multicolumn{6}{c}{cifar10} & \multicolumn{6}{c}{cifar100}    \\
         &\multicolumn{3}{c}{MAPE}
         &\multicolumn{3}{c}{RMSE}
         &\multicolumn{3}{c}{MAPE} 
         &\multicolumn{3}{c}{RMSE}\\
        \cmidrule(lr){2-4} \cmidrule(lr){5-7}  \cmidrule(lr){8-10}  \cmidrule(lr){11-13}  
        Epochs & 80 & 140 &200& 80 & 140 &200& 80 & 140 &200& 80 & 140 &200\\
        \midrule
         LC-BNN & 0.2978 & 0.3847 & 0.5879 & 0.1990 & 0.1963 & 0.2174 & 0.2843 & 0.2157 & 0.1867 & 0.2656 & 0.2155 & 0.1871 \\
         LC-PFN& 0.2274  & 0.3797  & 0.6008  & 0.1166  & 0.1519  & 0.1961 & 0.0557  & 0.0922  & 0.1505  & 0.0542  & 0.0834  & 0.1282   \\
        VRNN & 0.2396  & 0.2386  & 0.2378  & 0.1301  & 0.1371  & 0.1407 & 0.2138  & 0.2131  & 0.2113  & 0.1652  & 0.1668  & 0.1661    \\
         LSTM & 0.1851 & 0.1993 & 0.2093 & 0.1110 & 0.1092 & 0.1017 & 0.0502 & 0.0613 & \underline{0.0709} & 0.0526 & 0.0614 & 0.0648\\ 
         NODE &0.1655  & 0.1810  & 0.2039  & 0.1017  & 0.1026  & 0.0978   & 0.0705  & 0.0808  & 0.0823  & 0.0708  & 0.0788  & 0.0775   \\
         NSDE & \underline{0.1518}  & \underline{0.1555}  & \underline{0.1639}  & \underline{0.0956}  & \textbf{0.0913}  & \textbf{0.0839}  & \underline{0.0477}  & \underline{0.0597}  & 0.0723  & \underline{0.0492}  & \underline{0.0587}  & \underline{0.0647}   \\ 
         LC-GODE &\textbf{0.1487}  & \textbf{0.1536}  & \textbf{0.1629}  & \textbf{0.0953}  & \underline{0.0926}  & \underline{0.0860}   & \textbf{0.0460}  & \textbf{0.0571}  & \textbf{0.0630}  & \textbf{0.0481}  & \textbf{0.0577}  & \textbf{0.0598}  
  \\ 
        \bottomrule
    \end{tabular}
\end{table*}

\begin{table*}[ht]
    \centering
    \caption{ Elapsed time to train each epoch and perform inference on the test set. } \label{tab:rt}
  \addtolength{\tabcolsep}{-0.15em}
    \begin{tabular}{l|cccccccc}
        \toprule
          & \multicolumn{2}{c}{car} & \multicolumn{2}{c}{segment}  & \multicolumn{2}{c}{cifar10} & \multicolumn{2}{c}{cifar100}    \\
        \cmidrule(lr){2-3} \cmidrule(lr){4-5}  \cmidrule(lr){6-7} \cmidrule(lr){8-9}    
           & Train & Test& Train & Test& Train & Test& Train & Test \\
        \midrule 
          LC-BNN &0.0350 & 0.0183 & 0.0395 & 0.0558 & 0.0333 & 0.2833 & 0.0381 & 0.2814  \\
         LC-PFN &  -  & 9.6855 &  -  & 9.4088 &  -  & 109.8517 &  -  & 112.8367 \\
        VRNN &  0.4180 & 10.6489 & 0.4053 & 12.1125 & 1.7594 & 88.9615 & 1.5898 & 128.4061 \\
         LSTM & 0.8364 & 0.0005 & 0.8794 & 0.0005 & 6.5582 & 0.0043 & 6.7457 & 0.0049 \\ 
         NODE & 1.4698 & 0.1647 & 1.4684 & 0.1631 & 1.3543 & 0.1770 & 1.3206 & 0.1715 \\
         NSDE & 1.5931 & 0.3403 & 1.5817 & 0.3342 & 1.3347 & 0.3229 & 1.4389 & 0.3494 \\
           LC-GODE & 3.1136 & 0.8692 & 3.1476 & 0.8325 & 3.3043 & 0.7077 & 3.3879 & 0.7572 \\ 
        \bottomrule
    \end{tabular}
\end{table*}

Table~\ref{tab:extrapolate-loss} shows the extrapolation error of six baselines and LC-GODE for test loss curves originated from 2 tabular  tasks and 2 image classification tasks computed over three prediction lengths, observing 10 epochs. 
Table~\ref{tab:rt} shows the runtime (in seconds) of training one epoch and that of performing inference on the entire test set. For LC-PFN, the model is trained on synthetic curves drawn from a prior distribution and therefore no further training is needed, as long as the test learning curves are normalized according to the above description. Note that the inference time for Bayesian approaches, except for LC-BNN, is much greater than other approaches.

\end{document}